\documentclass[10pt,twocolumn,letterpaper]{article}

\usepackage[pagenumbers]{wacv} 

\newcommand\Tau{\mathcal{T}}
\usepackage{graphicx}
\usepackage[table,dvipsnames]{xcolor}
\usepackage{amsmath}
\usepackage{amssymb}
\usepackage{booktabs}
\usepackage{enumitem}
\usepackage{rotating}

\newtheorem{definition}{Definition}
\usepackage[symbol]{footmisc}
\usepackage{multicol}		
\usepackage{multirow}		
\usepackage[linesnumbered,ruled,vlined]{algorithm2e}
\usepackage{balance}
\usepackage{soul}
\usepackage{lipsum}

\newcommand{\bx}{\mathbf{x}}
\newcommand{\by}{\mathbf{y}}
\renewcommand{\paragraph}[1]{\vspace{1mm}\noindent\textbf{#1}}

\newlength\mylen

\SetKwFunction{Fun}{function}
\SetKwFunction{Fadapt}{adapt}
\SetKwFunction{Fevaluate}{evaluate}
\SetKwProg{Fn}{Function}{:}{}

\usepackage[pagebackref,breaklinks,colorlinks]{hyperref}

\usepackage[capitalize]{cleveref}
\crefname{section}{Sec.}{Secs.}
\Crefname{section}{Section}{Sections}
\Crefname{table}{Table}{Tables}
\crefname{table}{Tab.}{Tabs.}

\def\wacvPaperID{} 
\def\confName{WACV}
\def\confYear{2024}

\begin{document}

\title{Generalized Cross-domain Multi-label Few-shot Learning for Chest X-rays}

\author{
Aroof Aimen$^{1,2}$\stepcounter{footnote}\thanks{Work done during an internship at Wadhwani AI}
\quad Arsh Verma$^1$ \quad Makarand Tapaswi$^{1,3}$ \quad Narayanan C. Krishnan$^4$ \\
$^1$ Wadhwani AI \quad
$^2$ IIT Ropar \quad
$^3$ IIIT Hyderabad \quad
$^4$ IIT Palakkad
}

\maketitle

\begin{abstract}
Real world application of chest X-ray abnormality classification requires dealing with several challenges:
(i)~limited training data;
(ii)~training and evaluation sets that are derived from different domains; and
(iii)~classes that appear during training may have partial overlap with classes of interest during evaluation.
To address these challenges, we present an integrated framework called \emph{Generalized Cross-Domain Multi-Label Few-Shot Learning} (GenCDML-FSL), Fig.~\ref{fig:position-prob}.
The framework supports overlap in classes during training and evaluation, cross-domain transfer, adopts meta-learning to learn using few training samples, and assumes each chest X-ray image is either normal or associated with one or more abnormalities.
Furthermore, we propose \emph{Generalized Episodic Training} (GenET), a training strategy that equips models to operate with multiple challenges observed in the GenCDML-FSL scenario.
Comparisons with well-established methods such as transfer learning, hybrid transfer learning, and multi-label meta-learning on multiple datasets show the superiority of our approach.
\end{abstract}
\vspace{-3mm}

\section{Introduction}
\label{section:intro}

The recent Covid-19 pandemic has underscored the pivotal role of X-rays in healthcare, particularly in diagnosing and monitoring disease progression.
The surge in demand for X-ray analysis has highlighted a significant bottleneck: a shortage of radiologists to manually interpret these X-rays \cite{rsna_radiologists_shortage}.
Machine learning, with its potential to automatically detect abnormalities from chest X-rays, emerges as a promising solution to this challenge.
However, the dominant Machine Learning techniques, especially deep neural networks (DNNs), are data-hungry and require a vast amount of labeled data for effective performance.
The time-intensive nature of labeling X-rays manually further intensifies the pressure on an already strained healthcare system.

\begin{figure}[t]
\centering
\includegraphics[width=0.9\linewidth]{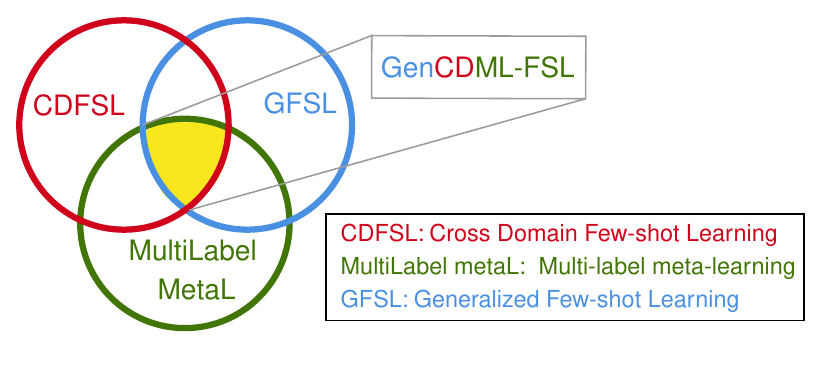}
\vspace{-2mm}
\caption{We propose a new scenario to overcome challenges of deploying real-world chest X-ray classification systems. Generalized Cross-domain Multi-label Few-shot Learning (GenCDML-FSL) lies at the intersection of Cross-domain FSL~\cite{das2021confess, wang2021cross}, Multi-label meta learning~\cite{alfassy2019laso,simon2022meta}, and Generalized FSL~\cite{gidaris2018dynamic,kukleva2021generalized}.
}
\label{fig:position-prob}
\vspace{-3mm}
\end{figure}

In response to this challenge, \textit{few-shot learning} has emerged as a subfield of machine learning that focuses on training DNNs with limited data while maintaining the ability to generalize to unseen images~\cite{wang2020generalizing,Jiang_2023_WACV,Wu_2023_WACV}.
Transfer learning or meta-learning are effective few shot learning approaches that
leverage the knowledge accumulated from related data to compensate for the scarcity of labeled examples for training~\cite{parnami2022learning,aimen2023adaptation}.
In \emph{transfer learning}, a model is trained on a large amount of labeled data (source data) followed by fine-tuning on a small number of samples from the domain of interest (target domain). The model is then evaluated on unseen examples from the target domain.
\emph{Meta-learning} (metaL)~\cite{finn2017model,aimen2023leveraging}, on the other hand, simulates the few-shot testing phase during the training process by organizing the training data into episodes. Each episode consists of a support set, which contains annotated data samples used for training the model, and a separate query set containing examples from the same classes for model evaluation. Multiple episodes are used during training to teach the model to adapt and generalize from limited data. 

\begin{figure}[t]
\centering
\includegraphics[width=0.9\linewidth]{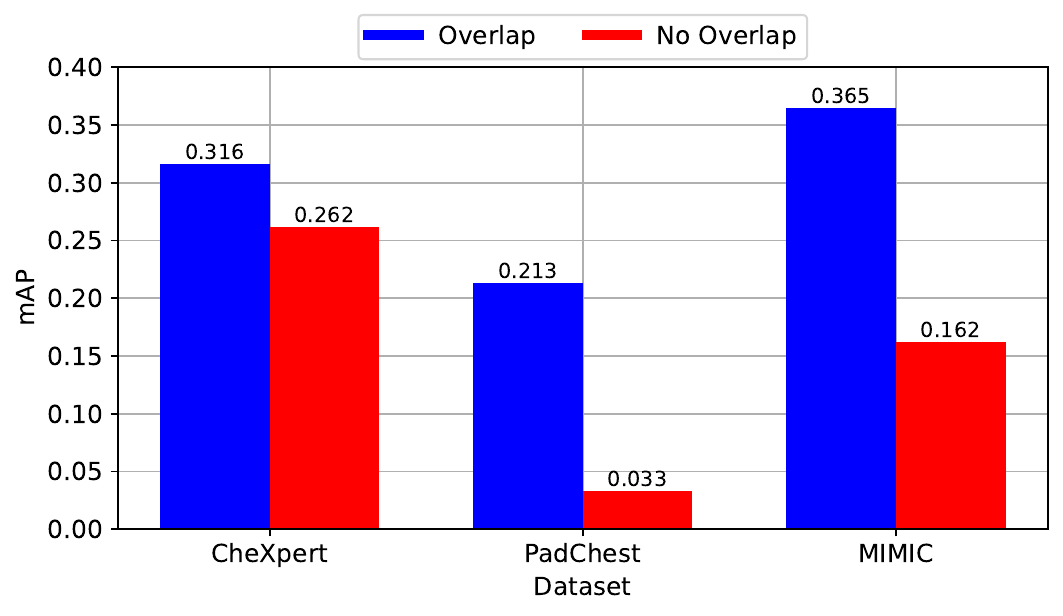}
\vspace{-4mm}
\caption{A significant performance drop is observed on non-overlapping labels when a model trained on NIH is evaluated post fine-tuning on three other datasets.}
\label{fig:over_non-over}
\vspace{-3mm}
\end{figure}

However, metaL for abnormality classification in X-ray images poses multiple challenges:
the X-rays utilized during training often differ in distribution from those encountered during testing (different machines, geography, and populations).
While this challenge may be solved through \textit{cross-domain few-shot learning} (CDFSL)~\cite{guo2020broader} methods, such approaches assume that the abnormalities (labels/ classes) in the training (meta-train) and testing (meta-test) sets are distinct -- and are therefore not applicable for our X-ray classification.
For example, a Resnet50 model trained on the NIH dataset~\cite{wang2017chestx, summers2019nih} has domain and label differences when fine-tuned and evaluated on CheXpert~\cite{irvin2019chexpert}.
The results shown in Fig.~\ref{fig:over_non-over}, indicate that the mean Average Precision (mAP) scores for test classes overlapping with NIH dataset are much higher compared to non-overlapping classes, suggesting a bias towards the classes encountered during pretraining.

Recognizing this gap, we introduce
\emph{Generalized Cross-domain Multi-label Few-shot Learning} (GenCDML-FSL), a new formulation at the intersection of multiple challenges (Fig.~\ref{fig:position-prob}).
(i)~The term \textit{Generalized} indicates (partial) overlap between the train and test labels~\cite{chao2016empirical}. This often introduces bias towards the overlapping (seen) classes, affecting the model's overall performance.
(ii)~\textit{Cross-domain} highlights the domain discrepancies between the train and evaluation data to which the model needs to adapt~\cite{guo2020broader}.
(iii)~\textit{Multi-label (ML)} applies since each X-ray image may show multiple abnormalities. In fact, the number of abnormalities also vary making this a challenging problem.
Finally,
(iv)~\textit{few-shot learning} refers to fine-tuning with limited data, which is a challenge in itself.

To address the challenges outlined in GenCDML-FSL, we propose a novel training pipeline called Generalized Episodic Training (GenET).
Each task within the GenET training pipeline consists of support, \emph{finetune}, and query sets, which are used for training, finetuning, and evaluating the model respectively.
The classes in the support and finetune sets may or may not overlap, simulating the evaluation conditions.
Furthermore, to introduce cross-domain differences, we apply different image augmentations to the support, finetune, and query samples.
Thus, GenET ensures that the model learns to adapt to new labels and adjust to different augmentations needed to handle the cross-domain aspect of GenCDML-FSL.

Our contributions can be summarized as follows:
\begin{enumerate}[topsep=0pt,itemsep=-1ex,partopsep=1ex,parsep=1ex]
\item We formulate a new problem setup for practical application of automatic chest X-ray classification (Sec.~\ref{sec:method:gencdml-fsl}).
GenCDML-FSL considers the inclusion of both seen and unseen classes during the test phase, a domain mismatch between the training and target data, limitations on the number of classes and samples per class, and a small validation set. 
\item  We present a novel generalized episodic training (GenET) pipeline (Sec.~\ref{sect:genETsection}) that has training, finetuning, and evaluation stages to mimic the test conditions.
\item Comparisons with transfer learning and vanilla episodic training highlight the superior performance of GenET in addressing GenCDML-FSL (Sec.~\ref{sec:experiments}).
\item  We also show that models trained using an episodic curriculum (ours or vanilla) are better calibrated compared to models trained using transfer learning.
\end{enumerate}

\section{Related Work}
Several recent works that leverage deep neural networks for abnormality classification in chest X-rays~\cite{verma2022can,ke2021chextransfer,jain2022deep,tang2020automated} presuppose access to extensive datasets. 
However, labelled datasets for real-world applications are often quite small.

\paragraph{Chest X-rays and few-shot learning.}
To address this constraint, there has been rising interest in few-shot learning (FSL) for X-rays ~\cite{paul2021discriminative,abdrakhmanov2022few,paul2020fast}.
These approaches assume consistent domains between training and evaluation sets or distinct label spaces across them~\cite{gong2023cross,cohen2020limits}.
In this paper, we introduce Generalized Cross-domain Multi-label Few-shot Learning (GenCDML-FSL) -- a practical framework at the intersection of three areas (see Fig.~\ref{fig:position-prob}) -- that relaxes these assumptions.
We now review works in these three areas.

\paragraph{Cross-domain Few-shot Learning (CDFSL).}
There are several works on CDFSL~\cite{DBLP:conf/icml/LiYZH19,setlur2021two,Broome_2023_WACV,Aich_2023_WACV} with the aim to generalize to distant domain classes in a few-shot learning setup.
A few examples are the introduction of an instance normalization and memorized and restitution module~\cite{xumemrein2022},
using an autoencoder to learn features by jointly reconstructing inputs and predicting labels~\cite{liang2021boosting},
or parametric adapters with residual connections~\cite{li2022cross}.
Contrastive learning approaches are also used,
specially in combination with feature selection and with a mixup module that uses a few samples of the target data for image diversity~\cite{das2021confess},
and feature disentanglement to reduce domain bias~\cite{fu2022generalized}.
A similar idea introduces an intermediate domain created by mixing source and target domain images to bridge the domain gap~\cite{zhuo2022tgdm,aimen2022adversarial}.
While prior works, including the ones that evaluate on chest X-ray datasets~\cite{das2021confess, wang2021cross}, operate in a multi-class setup, we go beyond and consider a multi-label setup with an overlapping train-test label space and domain discrepancies between training and testing.

\paragraph{Self-supervised learning (SSL)} is also adopted for CDFSL, and broadly for cross-domain chest X-rays classification~\cite{verma2022can, azizi2021big, vu2021medaug, sowrirajan2021moco, gazda2021self, reed2022self}.
Oh~\etal~\cite{oh2022understanding} demonstrate SSL pretraining excels when the target domain greatly differs from the source or has low few-shot difficulty.
They introduce two innovative pretraining schemes that enhance performance.
Yuan~\etal~\cite{yuan2022task} propose a domain-independent task-level SSL method that performs label-based instance-level supervision with task-level self-supervision using task consistency.

\paragraph{Multi-label meta-learning.}
The first work on multi-label FSL~\cite{alfassy2019laso} operates in the feature space on input pairs and their corresponding labels for sample synthesis. 
Simon~\etal~\cite{simon2022meta} improve this by extending the conventional methods of single-label FSL to the multi-label setting, and present a neural module for label count estimation using relational inference. This also serves as our baseline (ML-metaL).

To our best knowledge, there is only one work with multi-label metaL in chest X-rays~\cite{moukheiber2022few}.
This uses distribution calibration and ProtoNet in combination with geometric ensembles to identify uncommon labels.
Our work is different as we do not assume consistent train and test distributions (domains).

\paragraph{Generalized Few-shot Learning (GFSL)}
involves training models that are required to learn about rare/tail categories with few shots and simultaneously classify the sample among common/ head classes.

Chest X-ray datasets~\cite{summers2019nih, bustos2020padchest, irvin2019chexpert, johnson2019mimic} suffer from the problem of a long-tail distribution~\cite{zhou2021review}, which can make models susceptible to failure against rare and novel classes. Although previous work has explored
long-tail studies~\cite{galdran2021balanced, ju2021relational} for other medical imaging modalities, there is only fledgling work specifically for chest X-rays~\cite{holste2022long, holstecxr}.
Our work differs critically from Paul~\etal~\cite{paul2021generalized} and from Mahapatra~\etal~\cite{mahapatra2021medical}.
The former assumes the presence of textual reports that can provide semantic context, which may not always be the case in practice, while the latter does not use a multi-label setup.
Another direction that has been extensively studied to strengthen model performance, including that on rare classes, is to transfer from models pretrained on the same modality as the target dataset~\cite{sellergren2022simplified, matsoukas2022makes, reed2022self, sowrirajan2021moco} instead of ImageNet \cite{deng2009imagenet}, which is the default choice for most computer vision tasks.

Some methods adopted for GSFL outside of Chest X-rays include the use of an attention-based weight generator coupled with a ConvNet classifier redesigned as the cosine similarity function between feature representations and classification weight vectors aimed to learn stronger feature representations ~\cite{gidaris2018dynamic}. Ye~\etal~\cite{ye2021learning} introduced `CASTLE', a method that synthesizes calibrated few-shot classifiers while maintaining multi-class classifiers for head classes using a shared neural dictionary.
Another approach is a three-stage framework~\cite{kukleva2021generalized} for generalized and incremental few-shot learning that learns base classes, calibrates classifiers for novel classes while averting catastrophic forgetting, and ensures holistic classifier calibration. 
Our work differs from these studies as they assume that samples belong to one only label and there exists a consistent distribution between training and testing data.

\section{GenCDML-FSL Formulation}
\label{sec:method:gencdml-fsl}

\paragraph{Preliminaries.}
Consider an FSL setup where the train dataset is denoted as $Train$, validation as $Val$ and a test as $Test$.
These datasets consist of classes $C_{Train}$, $C_{Val}$ and $C_{Test}$ and come from domains $D_{Train}$, $D_{Val}$ and $D_{Test}$, respectively.
The train dataset $Train$ is used for training the model, the validation set $Val$ is utilized for hyperparameter tuning and model selection, and the test dataset $Test$ is employed for evaluating the model's performance.

In a meta-learning setup, tasks denoted as $T$, randomly sample some classes from $C_{Train}$, $C_{Val}$ or $C_{Test}$, adhering to a task distribution $P(\Tau)$.
Each task $T_i$ is an $N$-way $K$-shot learning challenge, where $N$ specifies the number of classes and $K$ denotes instances per class.
Note, each task $T_i$ comes from one of $Train$, $Val$, or $Test$ datasets.
Classically, a task (\eg~from $C_{Train}$) consists of a support set $S_i$ with samples drawn from classes $C^{S_i} \in C_{Train}$ and a query set $Q_i$ with different samples, but from the same classes, \ie~$C^{Q_i} = C^{S_i}$.

\begin{figure*}
\centering
\includegraphics[width=0.8\linewidth]{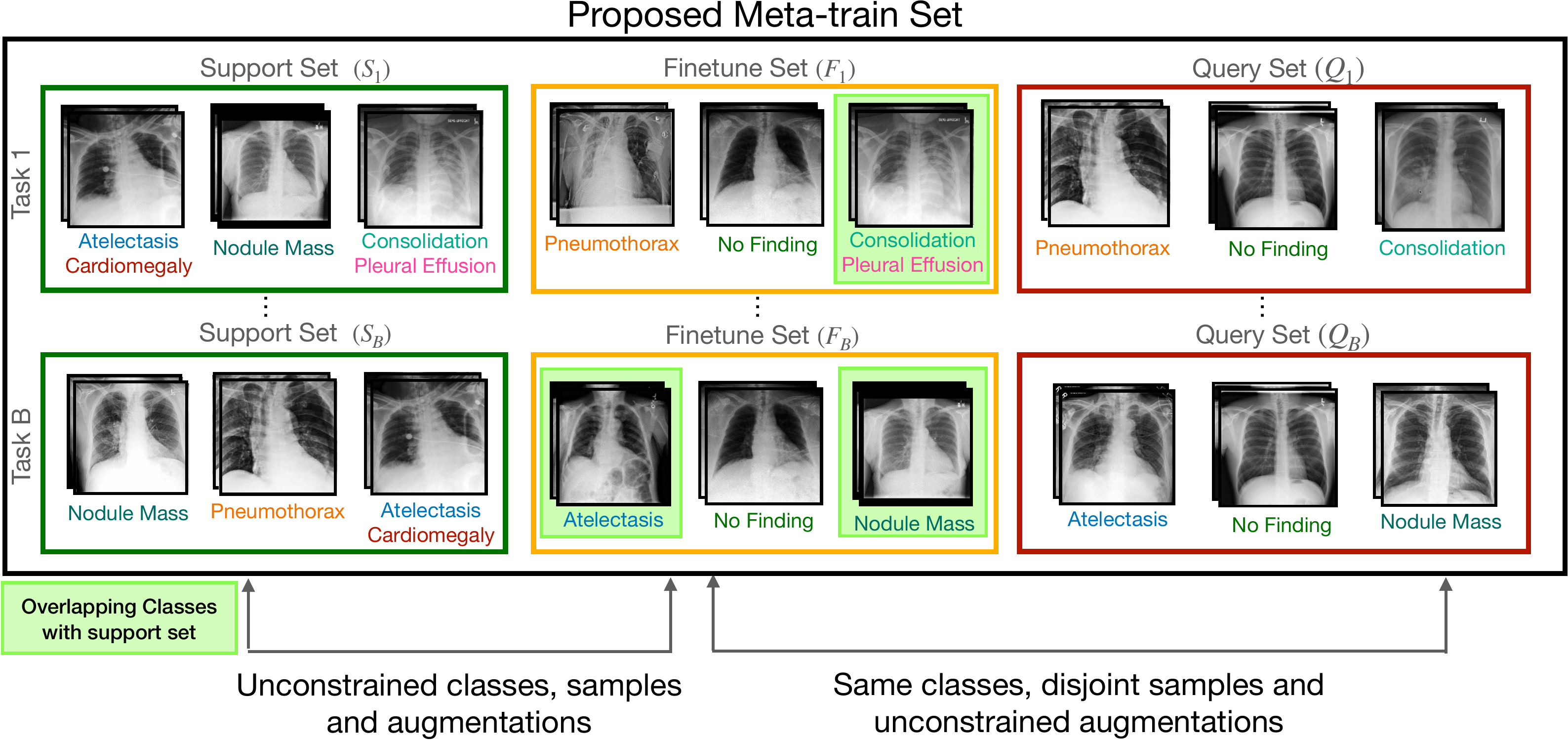}
\vspace{-2mm}
\caption{
A glimpse of the meta-train set for the Generalized Episodic Training (GenET) framework.
Each episode consists of $B$ tasks and each task consists of a support, finetune, and query set for model training, finetuning, and meta-updating respectively.
While support/finetune and support/query classes may be different, finetune/query sets share the same label space.}
\label{fig:genET}
\vspace{-3mm}
\end{figure*}

\subsection{GenCDML-FSL Problem}
\label{gen-problem}

In a classical meta-learning paradigm, the train $C_{Train}$, and test classes $C_{Test}$ are mutually exclusive.
However, in chest X-rays datasets, labels from the training set such as the NIH dataset~\cite{summers2019nih} (\eg~\emph{Cardiomegaly}, \emph{Atelectasis}, \emph{Pneumothorax}) may be present in test datasets like CheXpert~\cite{irvin2019chexpert}.
We refer to this setup that allows overlap between classes as \emph{Generalized}-FSL (G-FSL).

Furthermore, a domain disparity, \eg~arising from the country of data collection, between the train (\eg~NIH) and test (\eg~CheXpert) set is characterized as Cross-domain FSL (CD-FSL).
Different from multi-class FSL, our formulation allows images to be associated with multiple labels, introducing the Multi-label FSL (ML-FSL) paradigm.
Our formulation GenCDML-FSL integrates all three forms: G-FSL, CD-FSL, and ML-FSL.

\begin{definition}
We define GenCDML-FSL as a setup where:
(i)~train and validation have the same classes $C_{Train} = C_{Val}$;
(ii)~train and test have some overlapping labels $C_{Train} \cap C_{Test} \neq \varnothing$, the empty set; and
(iii)~the domains are assumed different: $D_{Train} \neq D_{Val} \neq D_{Test}$.
The paired samples in the train set may be written as
$Train = \{ \bx_i, \by_i \}_{i=1}^{|Train|}$.
Each chest X-ray image $\bx_i$ may be associated with one or more labels: $\by_i = \{y_i^c\}_{c=1}^{|C_{Train}|}$, where
$y_i^c \in \{0,1\}$ indicates presence or absence of the label $c$ in $\bx_i$.
$|C_{Train}|$ denotes the number of classes and $|Train|$ is the number of samples in the train set.
Similar multi-label definition can be adopted for the validation and test sets.
\label{def:def_gen_cross_multi_fsl}
\end{definition}

\paragraph{Multi-label episodic training.}
To accommodate the multi-label setup in episodic training, we relax the constraint on the number of shots per class to be equal to $K$ in a task while maintaining a fixed number of classes $N$.
For each task, we randomly select $N$ classes $(C^S)$ from a pool of available classes (\eg~from $(C_{Train}$) and disregard the remaining classes that are not selected, even if they are present in some samples.
For each selected class, we ensure that there are at least $K$ samples available for training.
Thus, the actual number of shots for each class range between $K$ and $N \times K$, depending on the presence of the label in other sampled instances of the same task.
This framework allows us to consider a multi-label setup for the selected classes of a task.
Note that across tasks, an image may be selected for its membership to different classes.

\paragraph{Validation set.}
As we focus on X-ray images (although GenCDML-FSL may be applied to other problems), typically $|C_{Train}|$ is small (few tens).
Thus, it is challenging to separate and allocate specific classes as part of the validation set ($Val$).
Furthermore, since the $Test$ set may not be available beforehand, a subset of samples from $Train$ are reserved as $Val$.
This means $C_{Train} = C_{Val}$, and while originally $D_{Train} = D_{Val}$, we break this domain similarity by applying differing image augmentations.

As a result, the $Val$ set does not truly represent the test set, introducing additional challenges.

\section{Generalized Episodic Training (GenET)}
\label{sect:genETsection}

To adapt episodic training to the generalized label space encompassing partially overlapping classes and to address train-test domain disparities, we introduce a novel training pipeline called Generalized Episodic Training (GenET).

Extending standard meta-learning, the GenET framework involves organizing each training episode (or task), denoted as $T_i$ into three distinct sets:
(i)~the support set $S_i = \{(\bx_k, \by^s_k)_{k=1}^K\}_{s=1}^N$ for training the model;
(ii)~a new \emph{finetune set} $F_i = \{ (\bx_p, \by^f_p)_{p=1}^{P}\}_{f=1}^N)$ for fine-tuning the model; and
(iii)~the query set $Q_i = \{(\bx_r, \by^q_r)_{r=1}^{R}\}_{q=1}^N)$ for evaluating the model's performance.
We use $\by_k^s$ to denote the multi-label vector $\by_k$ where category $s$ is present, \ie~$y_k^s = 1$, and others may or may not be $1$.

Similar nomenclature applies to $\by_p^f$ and $\by_r^q$.
$N$ represents classes in a task, and $K$, $P$, and $R$ represent the minimum number of samples belonging to each class in the support, finetune, and query sets, respectively.

During GenET, for a given task, classes within the support and query set may be partly overlapping, as shown in Fig.~\ref{fig:genET}.

However, the finetune and query sets have identical labels (\ie~$C^{F_i} = C^{Q_i}$), but the images in these sets are distinct from one another.

\paragraph{Learning procedure.}
The model parameters $\theta$ at iteration $t$ are adapted $U$ times on the support set $S_i$ through standard gradient descent on the support loss $L^s$, with a learning rate $\alpha$.
We denote the adapted model as
\begin{equation}
\label{eq:adapt}
\phi_i^U \leftarrow \theta^t - \alpha \nabla_{\theta^t} L^s(S_i; \theta^t) \, .
\end{equation}

The adapted model $\phi_i^U$ is fine-tuned $V$ times on $F_i$ using a learning rate $\beta$ and finetune loss $L^f$:
\begin{equation}
\label{eq:finetune}
\psi_i^V \leftarrow \phi_i^U - \beta \nabla_{\phi_i^U} L^f(F_i; \phi_i^U) \, .
\end{equation}

The fine-tuned model $\psi_i^V$ is subsequently evaluated on the query set $Q_i$ to obtain the query loss $L^q$.
The query loss, together with a learning rate $\gamma$ for all episodes in a batch of size $B$, is utilized to update the meta-model $\theta$
\begin{equation}
\label{eq:meta-up}
\theta^{t+1} \leftarrow \theta^t - \gamma \nabla_{\theta^t} \sum_{i=1}^B L^q(Q_i; \psi_i^Y) \, .
\end{equation}

To support the multi-label classification paradigm, we use binary cross-entropy loss over all samples and classes:
\begin{equation}
\label{eq:loss}
L = - \sum_k \sum_{j=1}^{C} y^j_k \log (\hat{y}^j_k) + (1 - y^j_k) \log (1-\hat{y}^j_k) \, ,
\end{equation}
where, $L$ stands for $L^s, L^f, L^q$ while $C$ corresponds to $C^S, C^F, C^Q$, the classes in the support, finetune and query set for the specific task.
$y^j_k$ denotes the true label for class $j$ of sample $\bx_k$ and $\hat{y}_k^j$ signifies the label probability predicted by the model.

Adapting the model on the support set $S_i$ and fine-tuning it on the finetune set $F_i$ trains the model for varying sets of classes.
Evaluating and fine-tuning the meta-model on the query set $Q_i$ makes the meta-model learn how to improve predictions even on unseen classes coming from different domains (augmentations).
Training the model in this manner promotes learning class- and domain-invariant features as the model exhibits better performance on both domain disparities and overlapping and non-overlapping classes.

Algorithm~\ref{alg:Generalized Episodic Training} summarizes the training procedure.
We provide a detailed explanation in the supplement.

\begin{algorithm}
\small
\SetAlgoLined
\KwIn{\\
\textit{Dataset:} $Train$ \\ 
\textit{Models:} Meta-model $\theta$, Base-model $\phi$\\
\textit{Learning rates:} $\alpha$, $\beta$, $\gamma$\\
\textit{Parameters:} Iterations $n_{iter}$, Batch-size $B$,\\
\hspace{50pt} Support adaptation-steps $U$, \\
\hspace{50pt} Finetune adaptation-steps $V$\\
}
\KwOut{Meta-model $\theta$}

\textbf{Initialization:} $\theta \leftarrow $ \text{Random Initialization} \\
\For{iter = 1, \ldots, $n_{iter}$}{ 
$\{T_i\} = \{S_i, F_i, Q_i\} \leftarrow$ SampleTasks($Train$, $B$) \\
\For{all $T_i$}{
$\phi_i^0 \leftarrow \theta$\\
$\phi_i^U=$ adapt$(\phi_i^0, S_i, \alpha, U)$ \\
$\psi_i^V=$ adapt$(\phi_i^U, F_i, \beta, V)$ \\ 
}  
$\theta \leftarrow \theta - \gamma \nabla_\theta \sum_{i=1}^B L^q (Q_i,\psi_i^V)$ \\
}
\textbf{Return} $\theta $ \\

\Fn{\Fadapt{$\phi_i^1, D_i, \mu, G$}}
{
\For{t=1, \ldots, G}{
$\phi_i^{t+1} \leftarrow \phi_i^t - \mu \nabla_{\phi_i^t} L(D_i; \phi_i^t)$} 
}
\textbf{\textit{Return} $\phi_i^{G}$ }
\caption{Generalized Episodic Training}
\label{alg:Generalized Episodic Training}
\end{algorithm}

\begin{figure}
\centering
\includegraphics[width=0.9\linewidth]{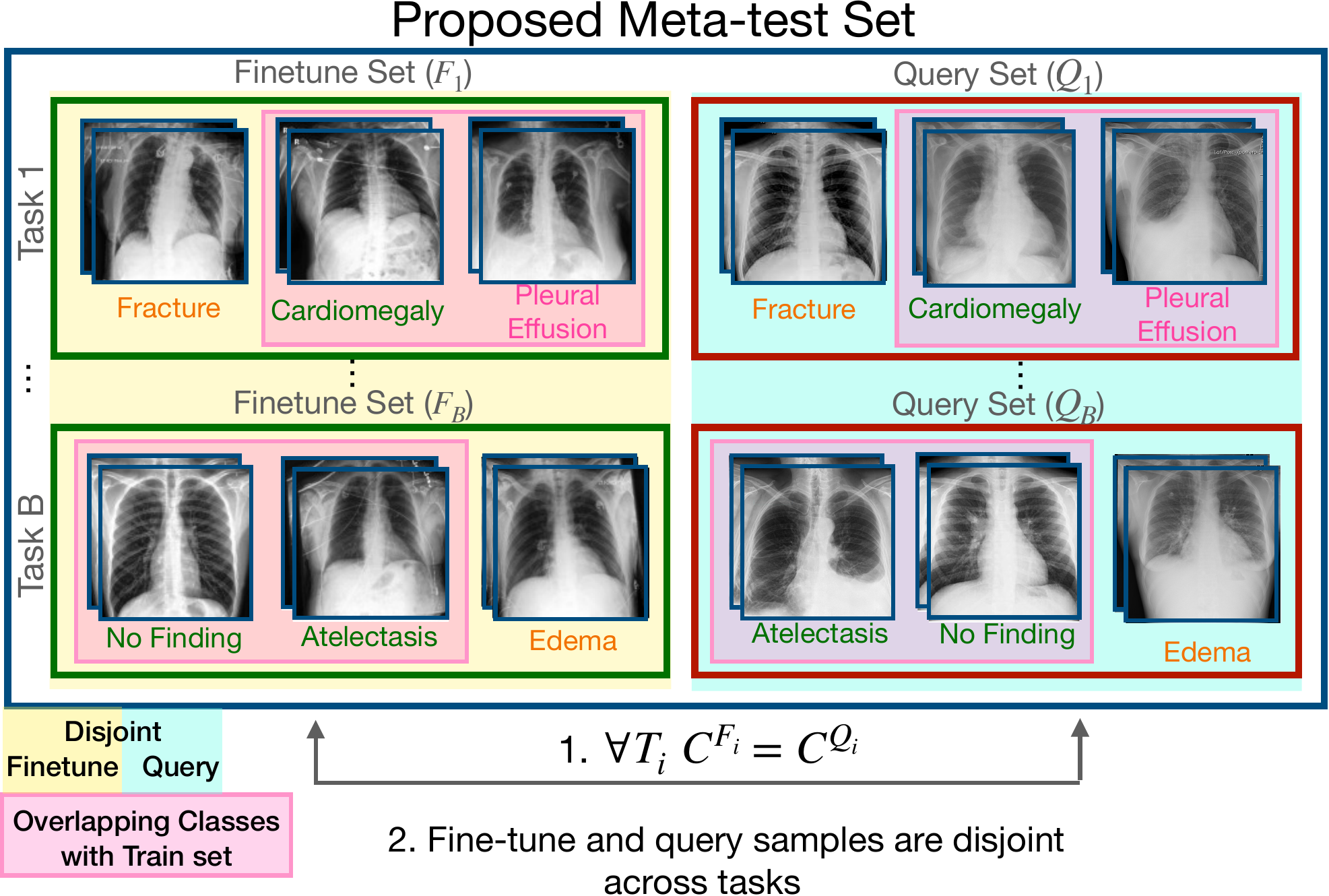}
\vspace{-2mm}
\caption{Proposed meta-test set has finetune and query samples that are disjoint from each other.
This enables a fair evaluation as only the finetuning set samples are labeled and used for adaptation in meta-testing and also other transfer learning frameworks.}
\label{fig:meta-test}
\end{figure}

\paragraph{Inference setup.}
\label{sect:target-split}
In conventional meta-learning, samples in the support set $S_i$ and query set $Q_i$ of task $T_i$ are disjoint during meta-testing.
However, it is possible for a sample from $Q_i$ to be present in the support set $S_j$ of another task $T_j$.
This assumes that the labels for such a sample are known, which is not true in real world applications.

To address this limitation, we propose a constrained meta-test set, where the intersection between all support sets and all query sets across tasks is empty, denoted as $S \cap Q = \varnothing$.
Thus, the samples from the support set, disjoint from the query set, are used for fine-tuning only for both evaluation frameworks meta-testing or simple transfer learning.
We thus refer to the support set as the finetune set (see Fig.~\ref{fig:meta-test}).
Additionally, similar to the proposed meta-train set (Fig.~\ref{fig:genET}), the classes in $F_i$ and $Q_i$ are same.

\section{Experiments}
\label{sec:experiments}

\begin{table}[t]
\small
\centering
\tabcolsep=0.15cm
\vspace{-2mm}
\begin{tabular}{l cccc}
\toprule
              & NIH & PadChest & CheXpert & MIMIC \\
\midrule
\# Labels     & 12      & 14       & 13       & 13        \\
NIH Overlap   & 12      & 12       & 8        & 8         \\
MIMIC Overlap & 8       & 9        & 13       & 13        \\
Test Samples  & 101437  & 145447   & 202135   & 341201    \\ 
\bottomrule
\end{tabular}
\vspace{-2mm}
\caption{Details of the datasets used in our experiments.
\emph{NIH (MIMIC) Overlap} indicates the number of labels that overlap with the dataset in the column.}
\vspace{-5mm}
\label{tab:dataset_details}
\end{table}

We conduct experiments to evaluate efficacy of the GenET algorithm for the Generalized Cross-domain Multi-label Few-shot learning paradigm using four popular chest X-ray datasets:
NIH~\cite{wang2017chestx,summers2019nih}, 
PadChest~\cite{bustos2020padchest},
CheXpert~\cite{irvin2019chexpert}, and
MIMIC~\cite{johnson2019mimic}.
Note that these datasets originate from varied geographical locations, are collected during different time periods, and have differences in their labeling processes and label sets themselves.
Therefore, training a model on one dataset (\eg~NIH or MIMIC in this paper) and testing it other datasets offers genuine cross-domain challenges with a few overlapping labels.
We augment labels by a \emph{Normal} category to indicate absence of all abnormalities.
Table~\ref{tab:dataset_details} summarizes some key aspects such as the overlapping labels.
More details are presented in the supplement.

\paragraph{Experimental setup.}
We perform pretraining on one dataset followed by fine-tuning or adaptation using few samples (240) of the test dataset in all experiments:
(i)~pretrain on NIH and evaluate on PadChest, CheXpert, or MIMIC;
(ii)~pretrain on MIMIC and evaluate on NIH, PadChest, or CheXpert.

\paragraph{Evaluation metrics.}
We evaluate multi-label classification performance in two ways.
A threshold-agnostic metric such as mean Average Precision (mAP) helps gain an overall sense of performance.

Deploying the model in the real-world requires choosing a threshold.
We report precision, recall, and F1 scores that have been used by previous works as well~\cite{paul2021discriminative,cores2022few}.
We report results for two thresholds.
The first is at 0.5, the theoretical optimal based on the binary cross-entropy loss.

A second threshold is assumed to be supplied by an oracle.
We pick 100 thresholds between 0 and 1 and choose the one that offers the best performance (F1 score) on the respective test dataset.
For a fair comparison, this approach is applied across all baselines as well.

\begin{table}[t]
\centering
\small
\tabcolsep=0.12cm
\begin{tabular}{l ccc ccc}
\toprule
\multirow{2}{*}{Methods} & \multicolumn{3}{c}{Source: NIH} & \multicolumn{3}{c}{Source: MIMIC}\\
 &  CX   &  PC   &  MIMIC   & CX & PC & NIH \\
\midrule
TL       & 0.2956  & 0.1871   & 0.2867  & 0.3658  & 0.1895  & 0.2040  \\
HTL      & 0.5616  & 0.5312   & 0.5331  & \textbf{0.6443}  & 0.5108  & 0.5615  \\
ML-metaL & 0.4278  & 0.3438   & 0.4324  & 0.4353  & 0.3648  & 0.3898  \\
MMAML    & 0.5332  & 0.5104   & 0.5175  & 0.6072  & 0.4871  & 0.4849  \\
\midrule
GenET  & \textbf{0.5773}  & \textbf{0.5340} & \textbf{0.5354} & 0.5687  & \textbf{0.5366}  & \textbf{0.6985} \\ 
\bottomrule
\end{tabular}
\vspace{-2mm}
\caption{Comparing all baselines against our approach GenET using \textbf{mean Average Precision (mAP)}.
CX is CheXpert, PC is PadChest.
GenET achieves highest score in 5 of 6 cases.}
\label{tab:mAP}
\vspace{-3mm}
\end{table}

\subsection{Baselines}
We compare GenET against multiple baselines:
(i)~transfer learning (TL)~\cite{guo2020broader},
(ii)~heterogenous transfer learning (HTL)~\cite{DBLP:conf/iclr/DhillonCRS20},
(iii)~multi-label MAML (MMAML)~\cite{finn2017model}, and
(iv)~a state-of-the-art multi-label meta-learning algorithm (ML-MetaL)~\cite{simon2022meta}.

\paragraph{Cross-domain approaches.}
Based on the findings from recent studies~\cite{chen2021meta,DBLP:conf/iclr/DhillonCRS20,chen2019closer}, transfer learning emerges as the leading baseline approach for cross-domain FSL in deep models.
We study two distinct transfer learning scenarios.

\paragraph{1. Transfer learning}
(TL) employs supervised fine-tuning followed by standard evaluation.
Here, the pretrained model is fine-tuned with few samples from the test domain and subsequently assessed on the main test set.
Specifically, in our experiments, we pretrain on the NIH/MIMIC datasets, fine-tune on 240 samples, and evaluate on the unseen samples of the three other datasets.

\paragraph{2. Hybrid transfer learning} (HTL)~\cite{DBLP:conf/iclr/DhillonCRS20} modifies the evaluation phase by structuring it into multiple tasks.
pretraining is performed in a standard supervised setup (similar to TL).
However, the evaluation is performed in an episodic manner and the dataset is split into finetune set for adaptation (240 samples) and the query set (rest of the samples) for evaluation.
Using the same number of adaptation samples (240) ensures a fair comparison as all methods are exposed to the same set of samples for fine-tuning.

\begin{table*}[t]
\small
\centering
\begin{tabular}{l l ccc ccc ccc}
\toprule
\tabcolsep=0.12cm
\multirow{2}{*}{\textbf{Th.}} &
\multirow{2}{*}{\textbf{Methods}} &
\multicolumn{3}{c}{\textbf{NIH $\rightarrow$ CheXpert}} &
\multicolumn{3}{c}{\textbf{NIH $\rightarrow$ PadChest}} &
\multicolumn{3}{c}{\textbf{NIH $\rightarrow$ MIMIC}} \\
&          & F1 Score & Precision & Recall & F1 Score & Precision & Recall & F1 Score & Precision & Recall \\
\midrule
\multirow{5}{*}{\rotatebox[origin=c]{90}{0.5}}
& TL       & 0.1818   & 0.3815    & 0.1426 & 0.0926   & 0.3794    & 0.0887 & 0.1544   & 0.3790    & 0.1369 \\
& HTL      & 0.3290   & 0.4417    & 0.3247 & 0.2345   & 0.4780    & 0.1882 & 0.3064   & 0.4984    & 0.2694 \\
& ML-metaL & 0.2082   & 0.2298    & 0.2667 & 0.0706   & 0.0752    & 0.1005 & 0.2126   & 0.2245    & 0.2738 \\
& MMAML    & \textbf{0.4067}   & 0.4600    & \textbf{0.4258} & \textbf{0.3683  } & 0.4355    & \textbf{0.3549} & \textbf{0.3904 }  & 0.4851    & \textbf{0.3785 }\\
& GenET    & 0.3666   & \textbf{0.4766}    & 0.3529 & 0.3327   & \textbf{0.4867 }   & 0.2885 & 0.3867   & \textbf{0.5152 }   & 0.3594 \\
\midrule
\multirow{5}{*}{\rotatebox[origin=c]{90}{Oracle}}
& TL       & 0.3336   & 0.2599    & 0.5093 & 0.2126   & 0.1825    & 0.3120 & 0.3225   & 0.2559    & 0.4637 \\
& HTL      & 0.5422   & \textbf{0.4838}    & 0.7627 & 0.4746   & \textbf{0.4326 }   & 0.6651 & 0.5097   & 0.4389    & 0.7680 \\
& ML-metaL & 0.4965   & 0.3872    & 0.8384 & 0.1788   & 0.1481    & 0.2883 & 0.3375   & 0.2983    & 0.5026 \\
& MMAML    & 0.5481   & 0.4452    & 0.8571 & 0.4897   & 0.3938    &\textbf{ 0.7593} & 0.5255   & 0.4137    & \textbf{0.8319} \\
& GenET    & \textbf{0.5803 }  & 0.4590   &\textbf{ 0.8898 }& \textbf{0.4997 }  & 0.4144    & 0.7356 & \textbf{0.5476 }  & \textbf{0.4509 }  & 0.8269 \\
\midrule
&
&
\multicolumn{3}{c}{\textbf{MIMIC $\rightarrow$ CheXpert}} &
\multicolumn{3}{c}{\textbf{MIMIC $\rightarrow$ PadChest}} &
\multicolumn{3}{c}{\textbf{MIMIC $\rightarrow$ NIH}} \\
&          & F1 Score & Precision & Recall & F1 Score & Precision & Recall & F1 Score & Precision & Recall \\
\midrule
\multirow{5}{*}{\rotatebox[origin=c]{90}{0.5}}
& TL       & 0.2930   & 0.5073    & 0.2575 & 0.1185   & 0.3107    & 0.1061 & 0.1611   & 0.2944    & 0.2930 \\
& HTL      & 0.3196   & 0.5221    & 0.2749 & 0.3085   & \textbf{0.4931}    & 0.2697 & 0.3526   & 0.5363    & 0.3016 \\
& ML-metaL & 0.1944   & 0.2128    & 0.2682 & 0.0466   & 0.0897    & 0.0712 & 0.0877   & 0.1555  & 0.1066 \\
& MMAML    & 0.4151   & 0.5358    & 0.3934 & 0.2617   & 0.3412    & 0.2511 & 0.3162   & 0.4274    & 0.2931 \\
& GenET    & \textbf{0.4203}   &\textbf{ 0.5473 }   & \textbf{0.4020} & \textbf{0.3303 }  & 0.4789    & \textbf{0.2956} & \textbf{0.5394}   & \textbf{0.6517 }   & \textbf{0.5033} \\
\midrule
\multirow{5}{*}{\rotatebox[origin=c]{90}{Oracle}}
& TL       & 0.3846   & 0.3343    & 0.4991 & 0.2046   & 0.1588    & 0.3823 & 0.2131   & 0.2414    & 0.5881 \\
& HTL      & 0.5727   & \textbf{0.5332}    & 0.6828 & 0.4347   & \textbf{0.4298 }   & 0.6636 & 0.4928   & 0.4713    & 0.7092 \\
& ML-metaL & 0.5297   & 0.3812    & \textbf{0.9476} & 0.3454   & 0.3198    & 0.5193 & 0.2767   & 0.2918   & 0.3734  \\
& MMAML    & \textbf{0.5850}   & 0.4853    & 0.7805 & 0.4931   & 0.3660    &\textbf{ 0.8603} & 0.4932   & 0.3884    & \textbf{0.7807} \\
& GenET    & 0.5733   & 0.4449    & 0.8922 & \textbf{0.4953 }  & 0.4074    & 0.7079 & \textbf{0.6363 }  &\textbf{ 0.6054 }   & 0.7092 \\
\bottomrule
\end{tabular}
\vspace{-2mm}
\caption{Comparing all baselines against GenET using threshold-based metrics such as F1 score, Precision, and Recall.
We report results for both thresholds: 0.5 and the best threshold denoted as Oracle.
Overall, GenET shows good performance across all 6 setups, while achieving highest F1 score for 5 of 6 tasks with oracle threshold.}
\vspace{-4mm}
\label{tab:main_threshold_table}
\end{table*}

\vspace{1mm}
We now discuss two multi-label meta-learning approaches as baselines.
Note that both approaches are not developed for the generalized label overlap setting.

\paragraph{3. Multi-label MAML} (MMAML)~\cite{finn2017model} is a simple extension of multi-class MAML for multi-label settings.
MMAML is similar to our approach GenET, except, MMAML does not include the finetune set $F_i$ as part of each episode.
Thus, it acts as a strong baseline to contrast against the technical novelty of GenET.

\paragraph{4. SotA multi-label meta-learning} (ML-MetaL)~\cite{simon2022meta} proposes a method that assumes each sample has two or more labels.
However, chest X-ray datasets have many images with a single label (\eg~28\% of NIH have a single abnormality while 54\% are labeled as normal (0 abnormalities)).
We exclude such samples from the NIH/MIMIC datasets during pretraining for consistency.

\subsection{Implementation Details}
We use a ResNet50 as the backbone for all experiments and resized all images to 128x128.

\paragraph{Learning rates (LR).}
For TL we set the LR as $10^{-4}$ based on best validation performance.
For HTL, we choose a LR of $10^{-4}$ for training and adaptation on meta-test finetune set following~\cite{finn2017model,chen2019closer}.
For MMAML and GenET, we set the support and query LR as $0.01$ and $0.001$ respectively. 
The meta-test finetune LR for GenET is the same as the meta-train support adaptation learning rate.
These values of LR are fixed for all approaches and datasets.

\paragraph{Episodic training details.}
For GenET, MMAML, and ML-MetaL,
the support adaptation steps are set to $U = 5$,
and the finetune steps are set to $V = 2$ for GenET.
It is important to note that the adaptation steps remain the same during meta-training and meta-testing and are consistent across all episodic experiments.
The intersection of support classes and finetune or query classes is chosen to be 0.3.

\paragraph{Batch size, Episode size, epochs.}
We set the batch size to 24 for non-episodic training and 1 for episodic training to reduce the computational burden~\cite{snell2017prototypical}.
Note that the batch used in episodic training corresponds to multiple tasks and cannot be directly compared to that used in non-episodic training which corresponds to individual samples.

We set the number of epochs for non-episodic training to 40.
To ensure fairness in terms of the amount of data seen by the model, we calculate the number of epochs for episodic training 
by $\text{NonEpisodicEpochs} = \text{40} \times \frac{\text{Total samples in dataset}}{\text{B} \times \text{Episode Size}}$.
$B=1$ is the batch size for episodic training.
The episode size is $N \times (K + P + R)$,
where $N=4$ represents the number of classes in a task,
$K=1, P=2, R=10$ represent the minimum number of samples (shots) belonging to each class in the support, finetune, and query sets respectively.
During evaluation, we set $P=1$.

\paragraph{Augmentations.}
The list of augmentations with specific parameters is: Horizontal Flip (p: 0.5), Vertical Flip (p: 0.2), Random Resized Crop (p: 0.5, height: 128, width: 128), Crop and Pad (p: 0.8, percent: [-0.3, 0.3]) and Rotation (p: 0.5). All other parameters are set to default as present in the Albumentations~\cite{buslaev2020albumentations} library. The augmentations are applied to each sample independently with two axes of freedom - strength of the augmentations and which subset of these are applied based on the associated probability parameters (i.e. p), which introduces stochasticity.

\paragraph{Meta-test finetune split.}
To ensure a fair process, we consider only 240 samples from the test set as annotated and use them across all methods for finetuning, meta-testing or transfer learning.
The finetune split is chosen to maximize the representation of all labels present in the test set.
In particular, we create 100 random splits, compute label distributions for the resulting finetune and test sets, and choose the one that minimizes label distribution distance.

\begin{figure*}
\centering
\includegraphics[width=\linewidth]{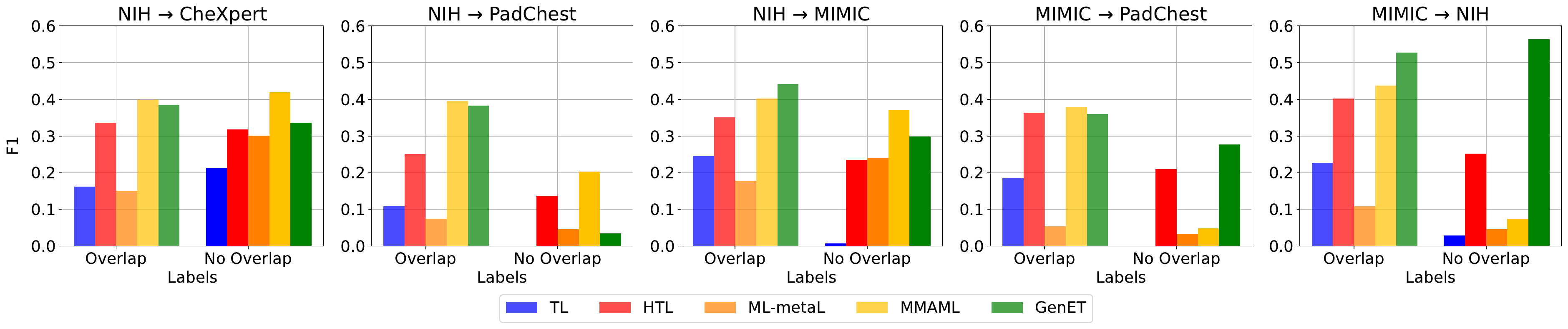} \\
\includegraphics[width=\linewidth]{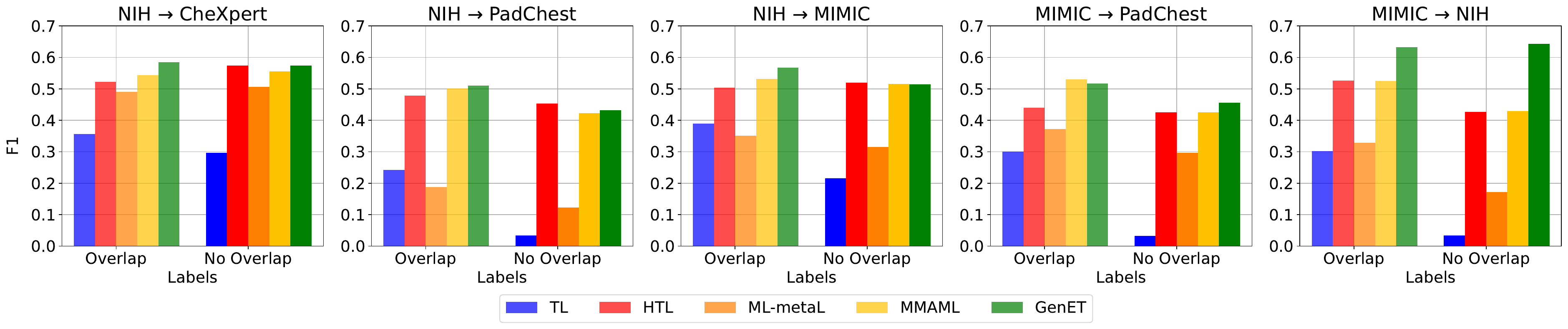}
\vspace{-7mm}
\caption{Mean F1 scores for overlapping (\emph{Overlap}) and non-overlapping (\emph{No Overlap}) classes.
\textbf{Top: 0.5 threshold} and \textbf{Bottom: oracle threshold}.
We ignore MIMIC $\rightarrow$ CheXpert as both datasets contain the same set of labels.
GenET outperforms baselines in majority of the cases and shows competitive performance with other meta-learning approaches.
HTL also outperforms TL in all instances.}
\vspace{-4mm}
\label{fig:overlap_analysis_both}
\end{figure*}

\subsection{GenET \vs~Baselines}

\paragraph{mAP evaluation.}
Table~\ref{tab:mAP} shows that GenET outperforms all baselines on mAP when using NIH as the source dataset.
We observe small, but consistent improvements of 2-4\% points on all three datasets.
Interestingly, transfer from MIMIC is more imbalanced, and GenET performs best in 2 of 3 cases.
From MIMIC to CX, the high score for HTL may be due to both datasets having same label space.

\paragraph{Evaluation at 0.5 threshold.}
Table~\ref{tab:main_threshold_table} reports standard classification metrics at the 0.5 threshold.
Analyzing F1 score, we see that GenET outperforms all baselines when the source dataset is MIMIC, while it comes a close second to MMAML when the source dataset is NIH.
Both TL and HTL achieve mediocre scores in this setting.

\paragraph{Evaluation at oracle threshold.}
Table \ref{tab:main_threshold_table} also reports results when using the oracle threshold.
Here, GenET outperforms all baselines on 5 of 6 cases when looking at oracle scores, and is a close second on MIMIC to CheXpert.

\paragraph{Impact of overlapping classes.}
Next, we analyze GenET's performance across overlapping and non-overlapping classes separately.
Fig.~\ref{fig:overlap_analysis_both} shows that GenET outperforms baselines in 6 of 10 instances each with overlapping labels and non-overlapping labels.
In particular, MIMIC to NIH shows large improvements with GenET both on overlapping and non-overlapping labels.
Additionally, note that non-overlapping classes have comparable performance to overlapping classes for GenET. 
This suggests that GenET enables the model to learn representations that are invariant to classes, validating our hypothesis that incorporating testing conditions into the training process through the finetune set enhances generalization.
For additional details, class-wise performance is reported in the supplement.

\paragraph{Comparison with SotA ML-metaL.}
Results presented in Table~\ref{tab:mAP} and Table~\ref{tab:main_threshold_table} show that GenET outperforms ML-metaL on all datasets and metrics.
The diminished performance of ML-metaL likely stems from its inability to manage domain discrepancies, a shared label space, and the stringent multi-label assumption.

\paragraph{GenET \vs~MMAML}
considers the impact of the finetune set to address the GenCDML-FSL problem.
Table~\ref{tab:mAP} shows GenET outperforms MMAML on the mAP metric.
In Table~\ref{tab:main_threshold_table}, F1 scores based on oracle threshold indicate that GenET outperforms MMAML in 5 of 6 cases, while at 0.5 threshold, the two approaches are closer.

\begin{table}[t]
\small
\centering
\tabcolsep=0.12cm
\begin{tabular}{l ccc ccc}
\toprule
\multirow{2}{*}{Methods} & \multicolumn{3}{c}{Source: NIH} & \multicolumn{3}{c}{Source: MIMIC} \\
         &  CX      &  PC      &  MIMIC   &   CX    &   PC   &  NIH  \\
\midrule
TL       &  0.5742  &  0.7791  &  0.6594 &  0.5078  &  0.6792  &  0.5722  \\ 
HTL      &  0.3898  &  0.5509  &  0.4018 &  0.5064  &  0.4855  &  0.4597  \\ 
ML-metaL &  0.3449  &  0.5136  &  0.3669 &  \textbf{0.3309}  &  0.4293  &  0.3883 \\ 
MMAML    &  0.3376  &  0.4669  &  0.3322 &  0.3982  &  \textbf{0.4079}  & \textbf{0.3834}  \\ 
GenET    &  \textbf{0.3280}    &  \textbf{0.4643}  &  \textbf{0.3292}  &  0.3531  &  0.4684  & 0.5081 \\ 
\bottomrule
\end{tabular}
\vspace{-4mm}
\caption{Comparing calibration of all baselines against GenET via \textbf{Expected Calibration Error (ECE)}.
Lower is better.
CX is CheXpert, PC is PadChest.
Models trained with meta-learning are better calibrated than those with transfer learning.
GenET performs well in 3 of 6 settings and is competitive on the others.}
\vspace{-3mm}
\label{tab:ece}
\end{table}

\subsection{Model Calibration}
Expected Calibration Error (ECE)~\cite{guo2017ece} measures the discrepancy between predicted probabilities and observed event frequencies. A well-calibrated model aligns predicted probabilities with actual likelihoods. 
We report the calibration error of all baselines and GenET in Table~\ref{tab:ece}.
The ECE for episodic training methods (MMAML, GenET) is notably lower than TL and HTL.
With NIH as the source dataset, GenET's ECE is the best.
Calibration is important when considering real-world applications and also hints towards the ease of choosing a threshold.

\section{Conclusion}
We introduced a new few-shot learning paradigm appropriate for real-world application of machine learning models to predict chest X-ray abnormalities.
\emph{Generalized Cross-domain Multi-label Few-shot learning} (GenCDML-FSL) encompasses overlapping and non-overlapping classes, domain disparities, and multi-label instances.
To address these challenges, we proposed \emph{Generalized Episodic Training} (GenET) that simulates challenges of GenCDML-FSL during the training process.
Through empirical validation, we demonstrated that adopting GenET significantly enhances the model's ability to learn class-invariant representations, outperforming transfer learning and other meta-learning baselines in majority of the cases across both overlapping and non-overlapping classes. 
We also observed that GenET and meta-learning in general results in better calibrated models, important for building user trust in the solution.

\paragraph{Acknowledgements.}
This work is made possible by the generous support of the American people through the United States Agency for International Development (USAID). The contents are the responsibility of Wadhwani AI and do not necessarily reflect the views of USAID or the United States Government.

\balance
{\small
\bibliographystyle{ieee_fullname}
\bibliography{longstrings,refs}
}

\appendix
\section*{Appendix}

\section{Generalized Episodic Training Algorithm}
\begin{figure*}[t]
\centering
\includegraphics[width=1\linewidth]{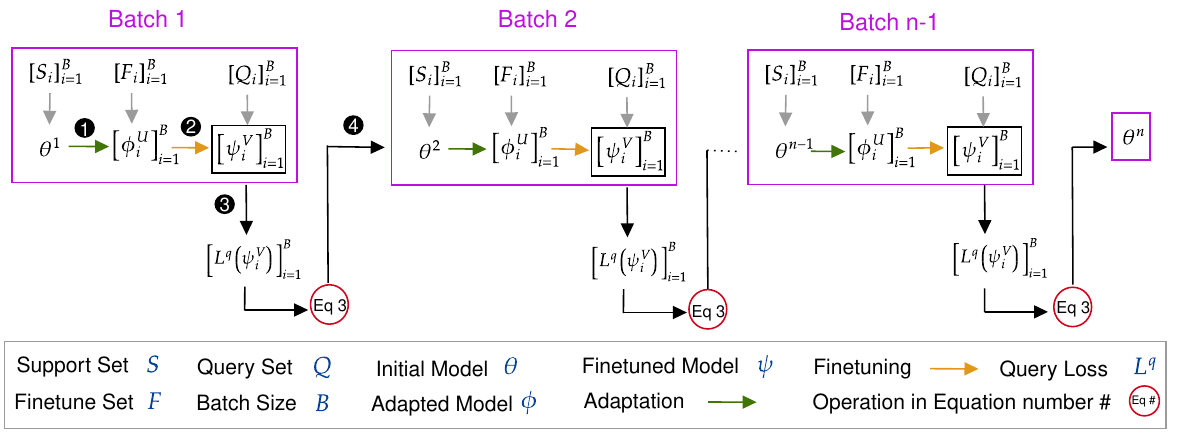}
\vspace{-6mm}
\caption{Computational graph of Generalized Episodic Training (GenET).
\textbf{Step 1: Adaptation}. Uses the support set $S_i$ to update meta-model parameters $\theta$ to $\phi$.
\textbf{Step 2: Fine-tuning}. Uses the finetune set $F_i$ to update $\phi$ to $\psi$.
\textbf{Step 3: Evaluation} is carried out over all $B$ tasks. We accumulate loss to update the meta-model.
\textbf{Step 4: Updates} the meta-model using the loss on the query set.
}
\vspace{8mm}
\label{fig:genET-computational}
\end{figure*}

Algorithm~\ref{alg:Generalized Episodic Training} and Fig.~\ref{fig:genET-computational} provide a concise summary of the Generalized Episodic Training approach.
The algorithm commences with the random initialization of the meta-model $\theta$, followed by the random sampling of a batch of tasks $\{T_i\}_{i=1}^B$ consisting of support, finetune, and query sets from the train dataset $Train$ (Alg.~1 lines 1-3).
The model $\theta$ is then adapted on the support set $S_i$ $U$ times (Fig.~\ref{fig:genET-computational} step 1), and subsequently on the finetune set $F_i$ for $V$ times (Fig.~\ref{fig:genET-computational} step 2).
Then the finetuned model $\psi_i^F$ is evaluated on the corresponding query set $Q_i$ (Fig.~\ref{fig:genET-computational} step 3).
Steps 1, 2, 3 of the figure correspond to lines 4-8 of the Alg.~1.
This entire process is performed for all tasks within the batch, and the query loss $L^q$ for all tasks in the batch is employed to update the meta-model $\theta$ (Fig.~\ref{fig:genET-computational} step 4, Alg.~1 line 9).

\section{Dataset Details}
\label{sec:datasets}

\paragraph{Overview.}
\label{sec:overview}
We leverage four publicly available datasets: NIH-CXR (NIH)~\cite{summers2019nih, wang2017chestx}, PadChest~\cite{bustos2020padchest}, CheXpert~\cite{irvin2019chexpert}, and MIMIC-CXR (MIMIC)~\cite{johnson2019mimic}.
These serve as the ideal choice for our problem formulation of Generalized Cross-domain Multi-label Few-shot learning (GenCDML-FSL) as they:
(i)~have partially overlapping labels that facilitate evaluation for generalization,
(ii)~originate from different geographical locations and time periods (CD),
(iii)~have diverse multi-label annotations,
and
(iv)~have a long-tail distribution which further motivates the few-shot learning problem.

Table~\ref{tab:data_sources} lists the details of our datasets that have been collected over decades from two different countries (including 3 states in the USA). 

\paragraph{Label mapping and overlap.}
\label{sec:label_map}
While preprocessing the datasets, we combined some of the labels present in our datasets based on inputs from multiple radiologists.

For NIH, we combined Infiltration, Consolidation and Pneumonia into ``Infiltration/Consolidation'' and Nodule and Mass into ``Nodule/Mass''.

The PadChest dataset contains 174 findings, 19 diagnoses and 104 anatomic locations. We extracted 14 important labels of interest by combining granular labels.

CheXpert and MIMIC use the same automated labeler to extract annotations. For these datasets, we combined Consolidation and Pneumonia into Infiltration/Consolidation and renamed Lung Lesion to Nodule/Mass.

Table~\ref{tab:labels} lists all the labels and the datasets which have this label.
We see that while labels like Atelectasis and Edema are present in all four datasets, Air Trapping is seen only in PadChest, Emphysema is seen only in NIH and PadChest, and Support Devices are annotated only in CheXpert and MIMIC. 
This complements Table~\ref{tab:dataset_details} that shows the partial overlap of labels between our pretraining datasets with all our evaluation datasets.

\begin{table}[b]
\small
\centering
\tabcolsep=0.065cm
\vspace{-3mm}
\begin{tabular}{l cccc}
\toprule
Label                      & NIH & PadChest & CheXpert & MIMIC \\
Total Count & 12 & 14 & 13 & 13 \\
\midrule
Infiltration/Consolidation & \checkmark & \checkmark & \checkmark & \checkmark \\
Emphysema                  & \checkmark & \checkmark & $\times$   & $\times$   \\
Edema                      & \checkmark & \checkmark & \checkmark & \checkmark \\
Atelectasis                & \checkmark & \checkmark & \checkmark & \checkmark \\
Nodule/Mass                & \checkmark & \checkmark & \checkmark & \checkmark \\
Pneumothorax               & \checkmark & \checkmark & \checkmark & \checkmark \\
Fibrosis                   & \checkmark & \checkmark & $\times$   & $\times$   \\
Cardiomegaly               & \checkmark & \checkmark & \checkmark & \checkmark \\
Hernia                     & \checkmark & \checkmark & $\times$   & $\times$   \\
Effusion                   & \checkmark & \checkmark & \checkmark & \checkmark \\
Pleural\_Thickening        & \checkmark & \checkmark & $\times$   & $\times$   \\
Pleural Other              & $\times$   & $\times$   & \checkmark & \checkmark \\
Fracture                   & $\times$   & \checkmark & \checkmark & \checkmark \\
Lung Opacity               & $\times$   & $\times$   & \checkmark & \checkmark \\
Enlarged Cardiomediastinum & $\times$   & $\times$   & \checkmark & \checkmark \\
Air\_Trapping              & $\times$   & \checkmark & $\times$   & $\times$   \\
Support Devices            & $\times$   & $\times$   & \checkmark & \checkmark \\
Normal                     & \checkmark & \checkmark & \checkmark & \checkmark \\ \bottomrule
\end{tabular}
\caption{Overview of labels in the datasets used in our study.
NIH~\cite{summers2019nih, wang2017chestx}, PadChest~\cite{bustos2020padchest}, 
CheXpert~\cite{irvin2019chexpert}, and MIMIC~\cite{johnson2019mimic},.
Note: CheXpert and MIMIC have the same set of labels.
}
\label{tab:labels}
\end{table}

\begin{table*}[]
\small
\centering
\vspace{-2mm}
\begin{tabular}{@{}ccccc@{}}
\toprule
& NIH         & PadChest      & CheXpert    & MIMIC       \\ \midrule
Source &
\begin{tabular}[c]{@{}c@{}}NIH Clinical Center, \\ MD, USA\end{tabular} &
\begin{tabular}[c]{@{}c@{}}San Juan Hospital, \\ Spain\end{tabular} &
\begin{tabular}[c]{@{}c@{}}Stanford Hospital, \\ CA, USA\end{tabular} &
\begin{tabular}[c]{@{}c@{}}Beth Israel Deaconess \\ Medical Center, MA, USA\end{tabular} \\
Time Period        & 1992 - 2015 & 2009 - 2017   & 2002 - 2017 & 2011 - 2016 \\
\# Images          & 112,120     & 160,868       & 224,316     & 377,110     \\
\# Patients        & 30,805      & 69,882        & 65,240      & 65,079      \\
\# Studies         & -           & 109,931       & -           & 227,827     \\
Reports Available? & No          & Yes (Spanish) & No          & Yes         \\
Annotation Process &
\begin{tabular}[c]{@{}c@{}}Automated Labeller, \\ Manual Validation\end{tabular} &
\begin{tabular}[c]{@{}c@{}}83\%: Automated Labeller, \\ 17\%: Manual\end{tabular} &
\begin{tabular}[c]{@{}c@{}}Automated Labeller, \\ Manual Validation\end{tabular} &
\begin{tabular}[c]{@{}c@{}}Automated Labeller, \\ Manual Validation\end{tabular} \\
\bottomrule
\end{tabular}
\vspace{-2mm}
\caption{Specifications of the datasets used in our experiments.}
\label{tab:data_sources}
\end{table*}
\section{Additional Results}

\subsection{Overlapping/ Non-overlapping Labels}
\label{sec:overlap_nonoverlap}
\begin{figure*}
\centering
\includegraphics[width=\linewidth]{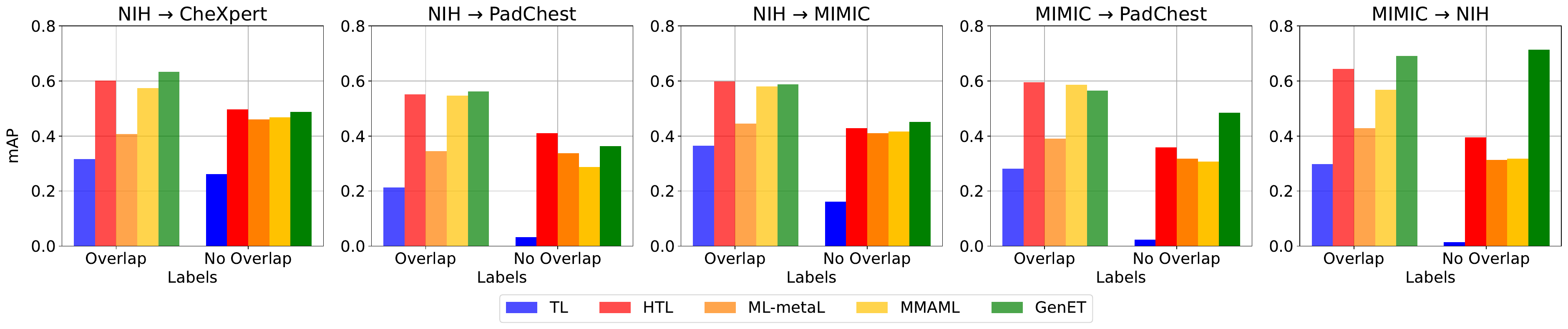}
\caption{\textbf{Mean Average Precision (mAP)} scores for overlapping (\emph{Overlap}) and non-overlapping (\emph{No Overlap}) classes.
We ignore MIMIC $\rightarrow$ CheXpert as both datasets contain the same set of classes.
GenET outperforms baselines in the majority of the cases and is competitive in others. It also compares favourably against other meta-learning approaches in almost all cases.
HTL also outperforms TL in all instances.}
\label{fig:mean_AP}
\end{figure*}

\begin{figure*}[t]
\centering
\begin{subfigure}{\linewidth}
\centering
\includegraphics[width=\linewidth]{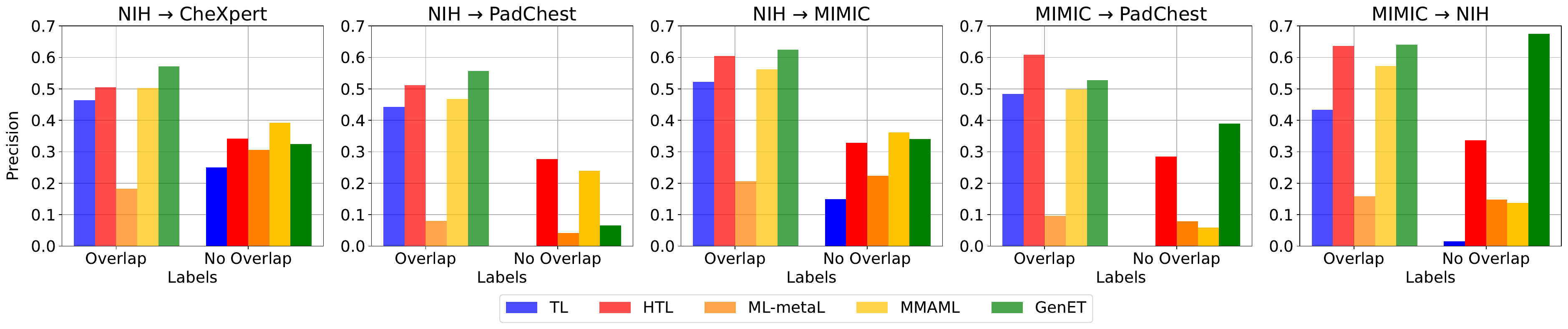}
\includegraphics[width=\linewidth]{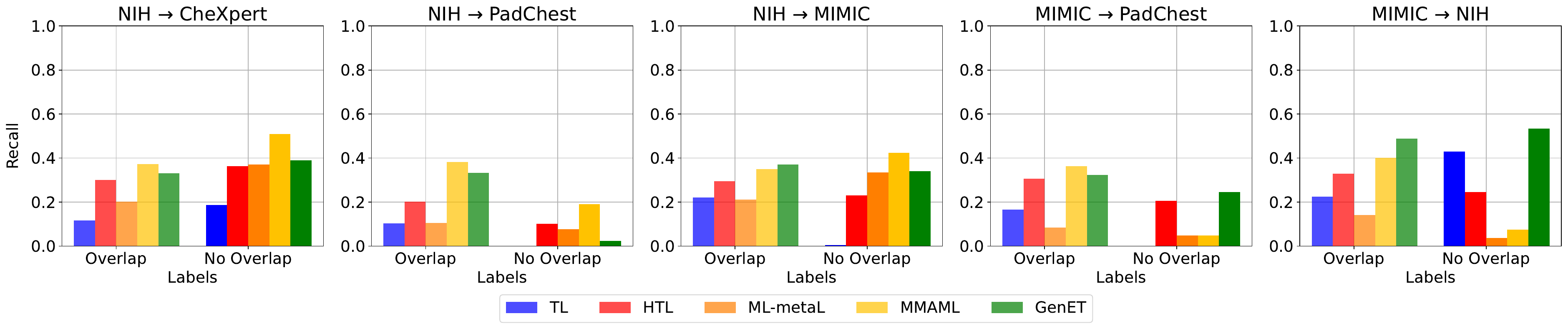}
\caption{Mean Precision and Recall at \textbf{0.5 threshold}.}
\end{subfigure}
\begin{subfigure}{\linewidth}
\centering
\includegraphics[width=\linewidth]{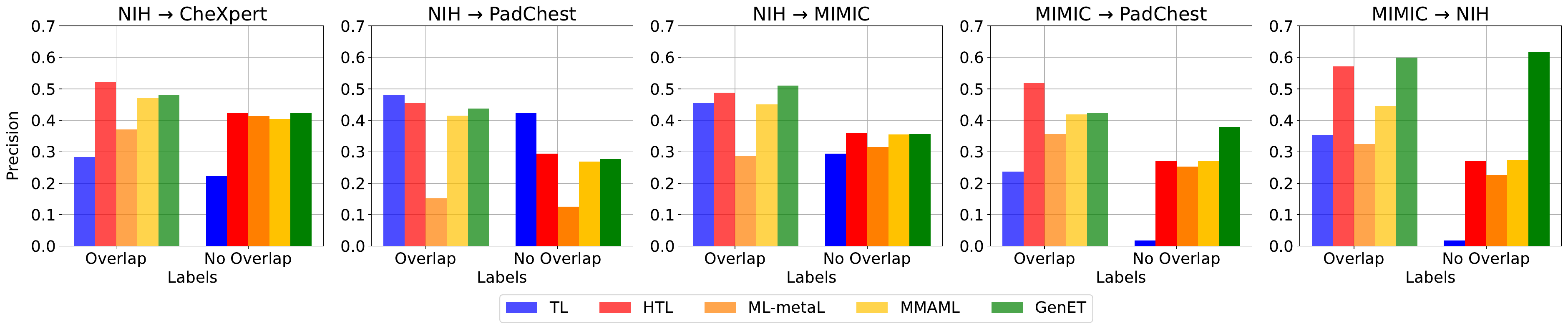}
\includegraphics[width=\linewidth]{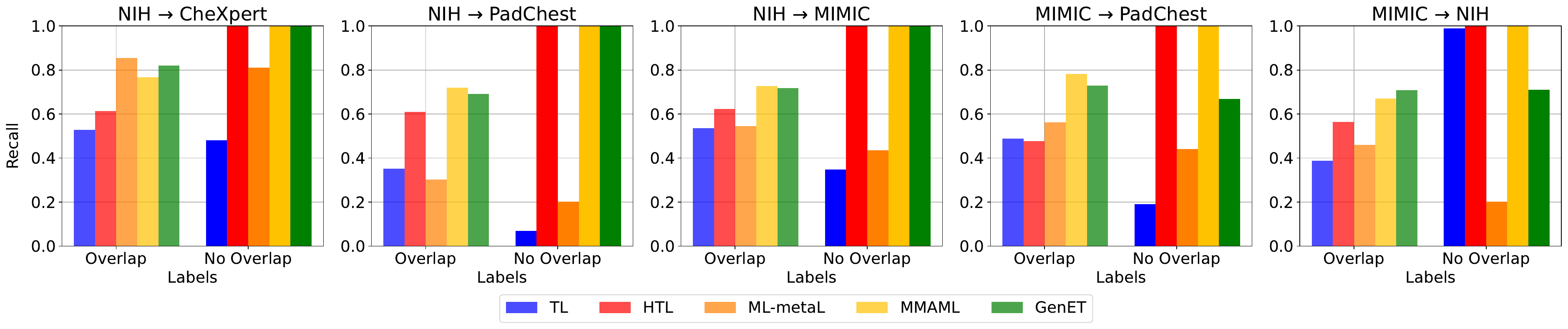}
\caption{Mean Precision and Recall at \textbf{oracle threshold}.}
\end{subfigure}
\caption{Mean \textbf{precision} and \textbf{recall} scores for overlapping (\emph{Overlap}) and non-overlapping (\emph{No Overlap}) classes at \textbf{threshold 0.5 (a, top)} and \textbf{oracle threshold (b, bottom)}.
GenET outperforms baselines in the majority of the cases.}
\label{fig:overlap_analysis_precision_recall_supp}
\end{figure*}

We see in Fig.~\ref{fig:mean_AP} that GenET consistently outperforms all other methods on mAP in cases where there is no overlap in the labels. 
The difference is particularly significant when models are trained on MIMIC.
Further, HTL always outperforms TL.
We also observe that in the no-overlap setting (non-overlapping labels), the TL performance is significantly worse than in all other methods.
On the overlapping labels, with the exception of ML-metaL, the gap between the methods is visibly lower. Even in this setting, GenET outperforms all other methods with the exception of the MIMIC~$\rightarrow$~PadChest. Again, HTL is a strong favourite over TL.

When we compare the precision and recall scores (refer Fig.~\ref{fig:overlap_analysis_precision_recall_supp}), we see that the trend is not as clear as mAP in Fig.~\ref{fig:mean_AP}.
GenET is better in 6/10 cases for the precision of overlapping labels, and is not much worse than the best method.
In particular, the GenET is worse for the MIMIC $\rightarrow$ PadChest scenario compared to HTL.
For the non-overlapping case, while the precision for GenET is the best only in 5/10 cases (5 transfer experiments, 2 thresholds), there is no single method that consistently beats GenET.

\subsection{Labelwise Results}
\label{sec:labelwise_results}
\begin{figure*}[p]
\centering
\includegraphics[width=\linewidth]{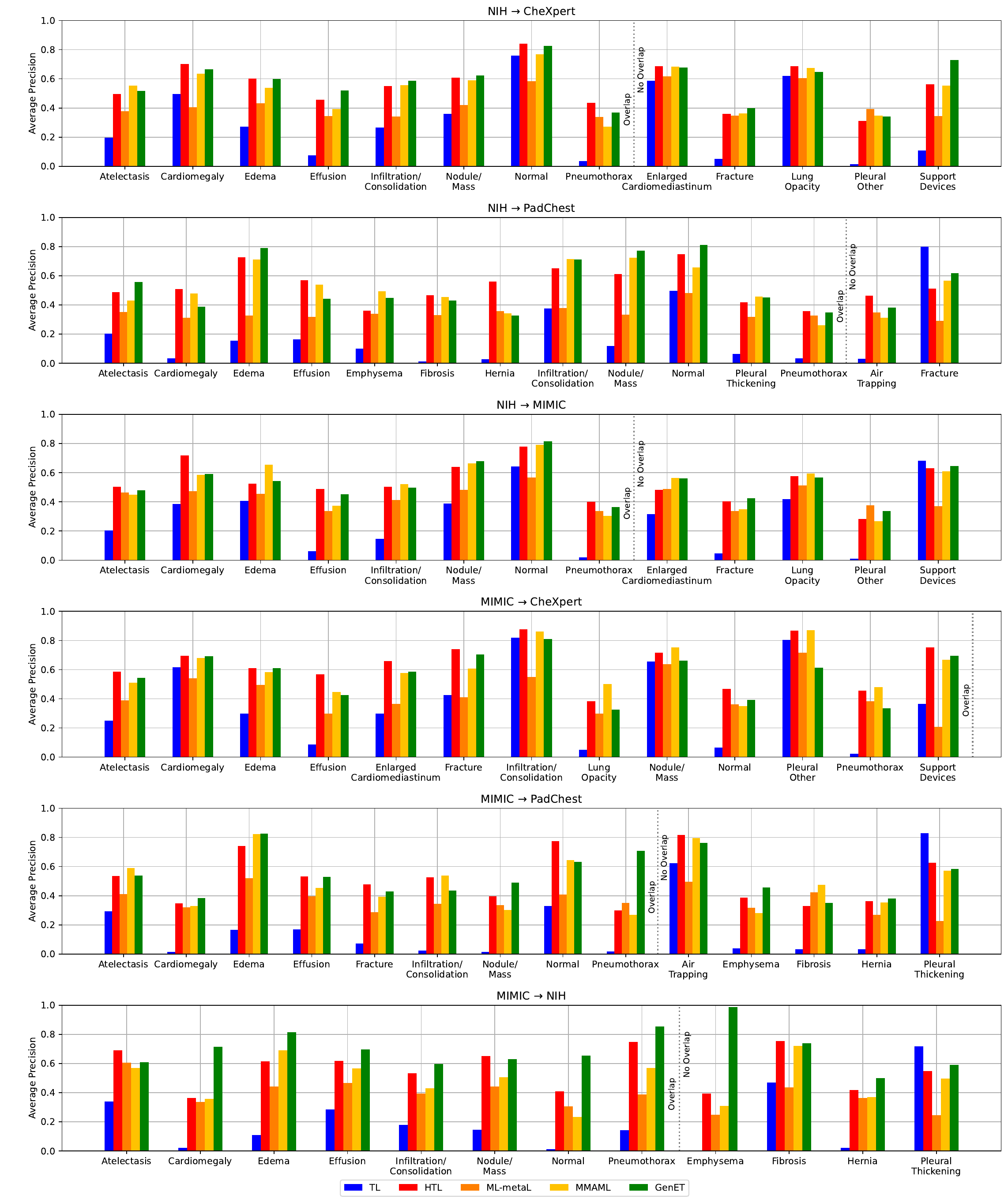}
\caption{\textbf{Label-wise average precision} (AP) scores for models trained on NIH (MIMIC) and evaluated on CheXpert, PadChest, and MIMIC (NIH).}
\label{fig:labelwise_AP}
\end{figure*}

\begin{figure*}
\centering
\includegraphics[width=\linewidth]{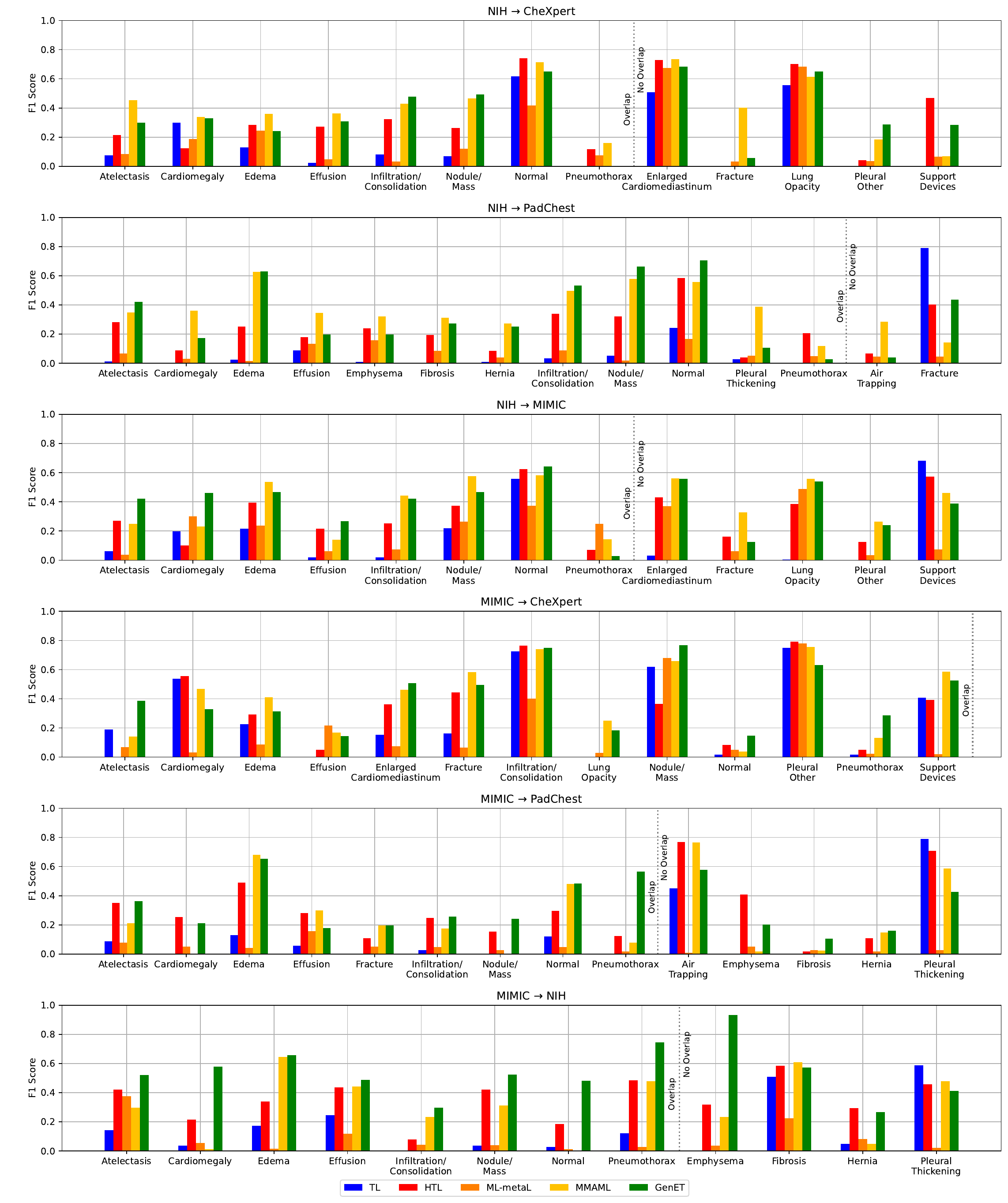}
\caption{\textbf{Label-wise F1 scores}  for models trained on NIH (MIMIC) and evaluated on CheXpert, PadChest, and MIMIC (NIH) when \textbf{threshold is 0.5}.}
\label{fig:labelwise_f1_05}
\end{figure*}

\begin{figure*}
\centering
\includegraphics[width=\linewidth]{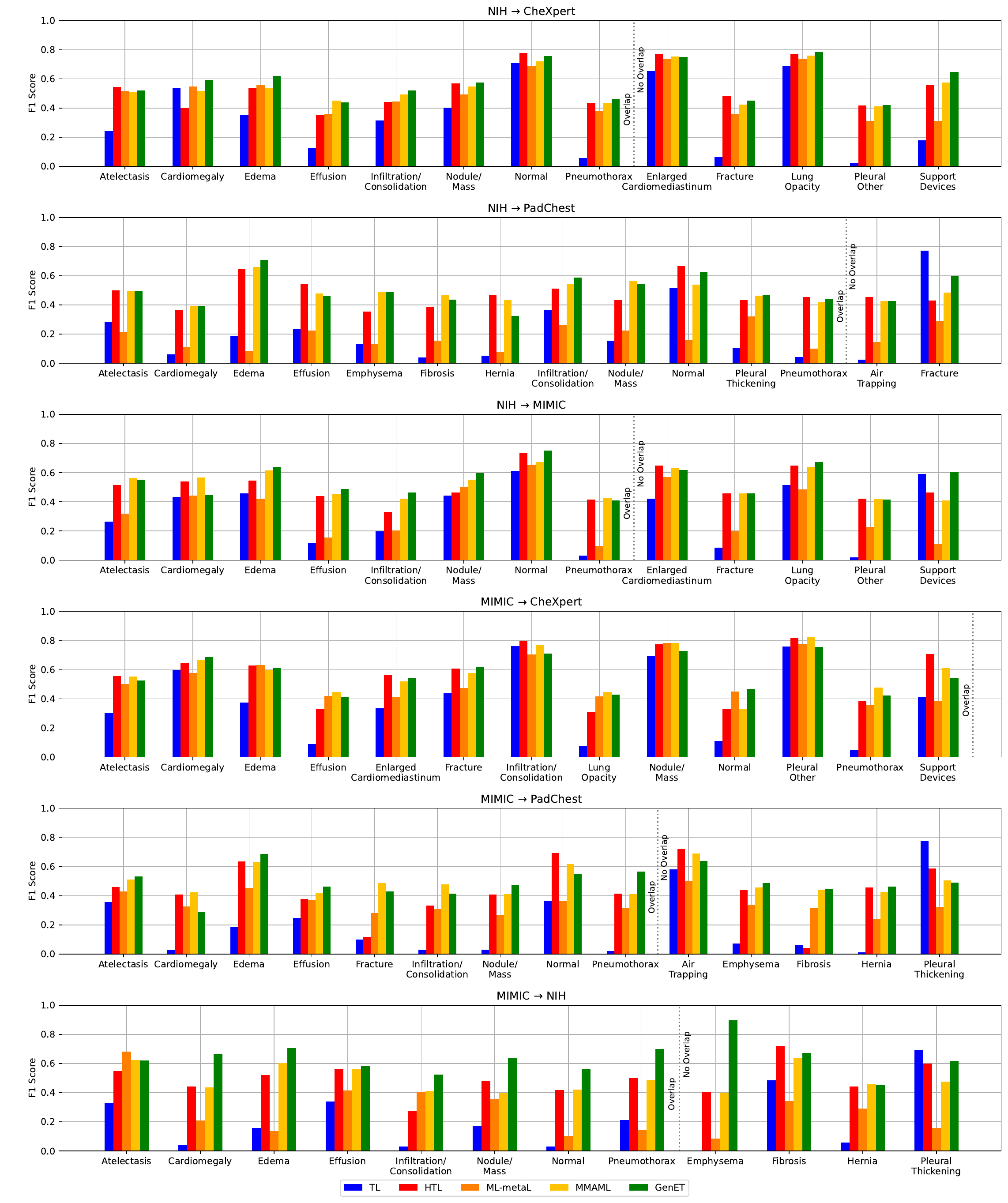}
\caption{\textbf{Label-wise F1 scores} for models trained on NIH (MIMIC) and evaluated on CheXpert, PadChest, and MIMIC (NIH) when \textbf{threshold is obtained from an oracle}.}
\label{fig:labelwise_f1_oracle}
\end{figure*}

We evaluate GenET against TL, HTL, ML-metaL, and MMAML for different labels across multiple datasets.
We report the AP (refer Fig.~\ref{fig:labelwise_AP}) and F1 scores at both 0.5 and a threshold from the oracle (refer~Fig.~\ref{fig:labelwise_f1_05}, \ref{fig:labelwise_f1_oracle}).
We partition the labels on the x-axis into overlapping and non-overlapping labels.
Note that for the MIMIC~$\rightarrow$~CheXpert experiment, all labels are the same. 

Looking at the AP values in Fig.~\ref{fig:labelwise_AP}, we observe a similar trend as in Sec.~\ref{sec:overlap_nonoverlap} that GenET generally outperforms all other methods on each label, which validates the merit of the method as the mean performance isn't getting biased by a major difference on just a few labels. This is the case for both overlapping and non-overlapping labels.
Similarly, Fig.~\ref{fig:labelwise_f1_05}, \ref{fig:labelwise_f1_oracle} also present the F1 scores label for each label, expanding on the results presented in Fig.~\ref{fig:overlap_analysis_both}.

\end{document}